%% file: main.tex
\PassOptionsToPackage{table}{xcolor}   
\documentclass{zlab}

\usepackage[T1]{fontenc}
\usepackage{lmodern}

\usepackage{xspace}
\usepackage{multirow}

\usepackage{graphicx,amsmath,amssymb,hyperref}
\usepackage{subcaption}
\usepackage{longtable}
\usepackage{listings}
\usepackage{tcolorbox}
\tcbuselibrary{listings,skins,breakable}
\usepackage{adjustbox}
\usepackage[authoryear,round]{natbib}
\usepackage{wrapfig}
\usepackage{placeins}
\usepackage{helvet}
\usepackage{etoc}
\usepackage[hang,flushmargin]{footmisc}

\usepackage{float}
\usepackage{array}
\usepackage[all]{hypcap}
\setlist[itemize]{itemsep=0.1pt}

\definecolor{promptbg}{HTML}{F9F9F9}
\tcbset{
    promptstyle/.style={
        enhanced,
        breakable,
        listing only,
        colback=promptbg,
        boxrule=0.9pt,
        left=10pt, right=10pt, top=8pt, bottom=8pt,
        listing options={
            language={},
            basicstyle=\fontsize{8}{9.6}\selectfont\sffamily,
            breaklines=true,
            breakatwhitespace=false,
            breakindent=0pt,
            columns=fullflexible,
            keepspaces=true,
            showstringspaces=false,
            frame=none,
            upquote=true,
            xleftmargin=0pt,
            xrightmargin=0pt,
            aboveskip=0pt,
            belowskip=0pt,
        },
        fonttitle=\sffamily\fontseries{bx}\selectfont,
        toptitle=5pt,
        bottomtitle=5pt,
        lefttitle=10pt,
        boxsep=0pt,
        sharp corners
    }
}

\definecolor{xstackcolor}{HTML}{1976D2}
\definecolor{linkred}{HTML}{ED1C24}

\hypersetup{
  colorlinks=true,
  urlcolor=xstackcolor,
  linkcolor=linkred,
  citecolor=xstackcolor,
  filecolor=black,
  pdfborder={0 0 0}
}

\usepackage{fontawesome5}   
\newcommand{\iconfallback}[1]{\raisebox{0.1pt}{\scriptsize\texttt{#1}}\xspace}
\newcommand{\github}{%
  \IfFileExists{figures/logos/github_logo.pdf}
    {\raisebox{-1.3pt}{\includegraphics[height=1.05em]{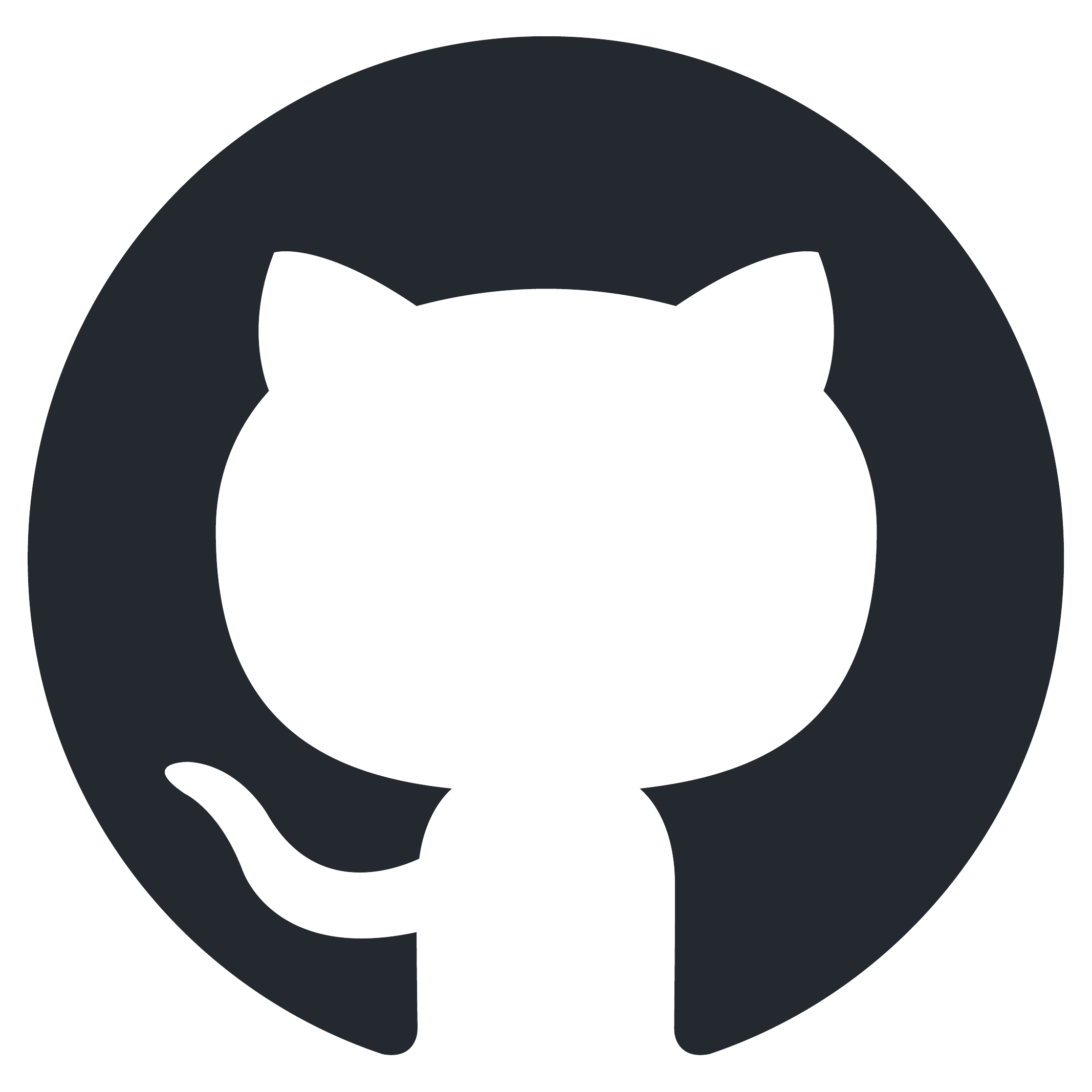}}\xspace}
    {\iconfallback{GH}}%
}
\newcommand{\hf}{%
  \IfFileExists{figures/logos/huggingface_logo.pdf}
    {\raisebox{-1.3pt}{\includegraphics[height=1.05em]{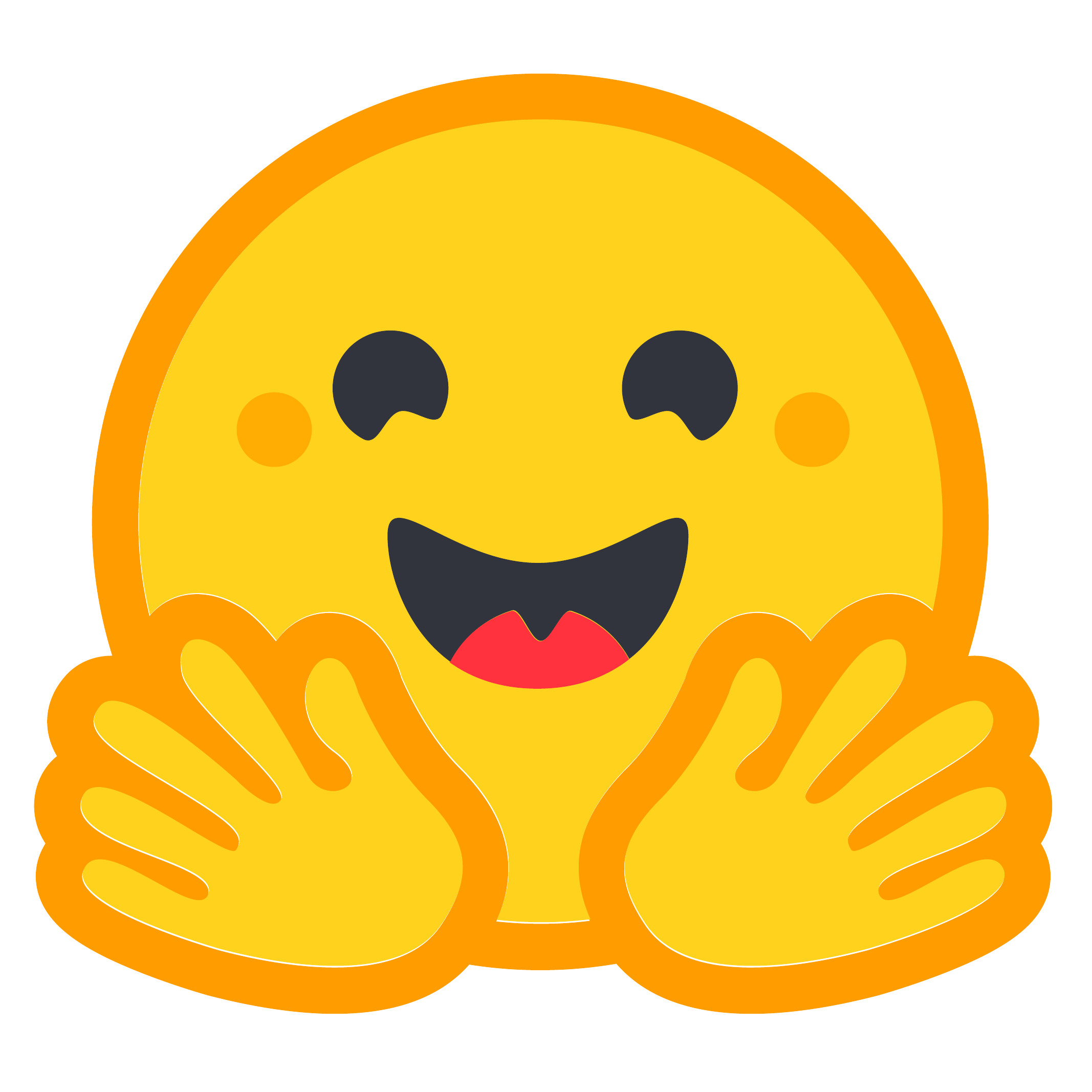}}\xspace}
    {\iconfallback{HF}}%
}
\newcommand{\websiteicon}{\raisebox{-0.6pt}{\textcolor{xstackcolor}{\faGlobe}}\xspace}
\newcommand{\linkstablestyleone}{%
    \begin{center}
        \normalsize
        \hf\,\href{https://huggingface.co/datasets/zlab-princeton/XStack}{\textbf{Dataset}}
        \qquad
        \github\,\href{https://github.com/zlab-princeton/XStack}{\textbf{Code}}
        \qquad
        \websiteicon\,\href{https://zlab-princeton.github.io/XStack/}{\textbf{Website}}
    \end{center}%
}

\newcommand{\bench}{XStack\xspace}

\input{figures/numbers.tex}

\definecolor{passgreen}{HTML}{1F8A5B}
\definecolor{failred}{HTML}{B9432D}
\newcommand{\file}[1]{\texttt{#1}}
\newcommand{\debugpar}{}

\makeatletter
\fancypagestyle{firststyle}{

  \fancyhead[L]{}
  \fancyhead[C]{}
  \fancyhead[R]{}
  \fancyfoot[L]{{\fontsize{8.5}{10}\selectfont \the\correspondingauthor}}
  \fancyfoot[R]{}
}
\makeatother

\DeclareTextFontCommand{\textbf}{\bfseries}

\title{%
  {\fontsize{16}{22}\selectfont XStack}{\fontsize{14}{20}\selectfont : A Benchmark for ML Repo Migration\par}
  \vspace{-0.5cm}
}

\author{
    \vspace{.2cm}
    \parbox{\textwidth}{\centering
        Linrong Cai \quad
        Taiming Lu \quad
        Haozhe Chen \quad
        Zhuang Liu
    }
    \\
    \vspace{-0.3cm}
    {\normalfont\fontsize{11}{15}\selectfont {Princeton University}\hspace{.1cm}}
        \vspace{.3cm}
}

\makeatletter
\renewcommand{\abscontent}{}%
\makeatother

\newenvironment{abstractblock}{%
  {\centering\large\bfseries Abstract\par}
  \vspace{0.1em}
  \begin{list}{}{%
      \setlength{\leftmargin}{2em}
      \setlength{\rightmargin}{2em}
      \setlength{\topsep}{0pt}
      \setlength{\parsep}{0pt}
  }
  \item[]
}{%
  \end{list}
  \par\normalfont\vspace{1em}
}

\begin{document}

\begingroup
\makeatletter
\let\raggedright\centering
\makeatother

\maketitle
\endgroup

\newcommand{\abstractcontent}{

\vspace{0.2cm}
ML code should run where researchers choose, not where it was first written. Today, machine learning code mainly lives in two ecosystems, PyTorch on NVIDIA GPUs and JAX on Google TPUs. Migrating repositories between them means building and testing across frameworks and hardware. This migration is (1) a practical need, since it unlocks different hardware for researchers, and (2) a natural benchmark: it involves different frameworks and hardware, covers whole repositories, and is long-horizon and challenging. We introduce \textbf{XStack}, a benchmark that evaluates agents on migrating ML repositories between these two ecosystems. Each agent receives a PyTorch repository and a clean TPU machine and must deliver a JAX repository. Its output passes if it meets our grading criteria, based on the source repository's published result,  authenticated and calibrated by us. Migration by agents remains unreliable: Claude~Opus~5 passes 75\%, GPT-5.6-sol 45\%, and every other agent at most 40\%. XStack is a step toward freeing ML code from its ecosystem. We release XStack to make progress toward reliable migration across frameworks and hardware measurable.
}

\newcommand{\teaserfigure}[2]{
    \begin{figure}[#1]
        \vspace{-0.2cm}
        \centering
        \includegraphics[width=0.9\linewidth]{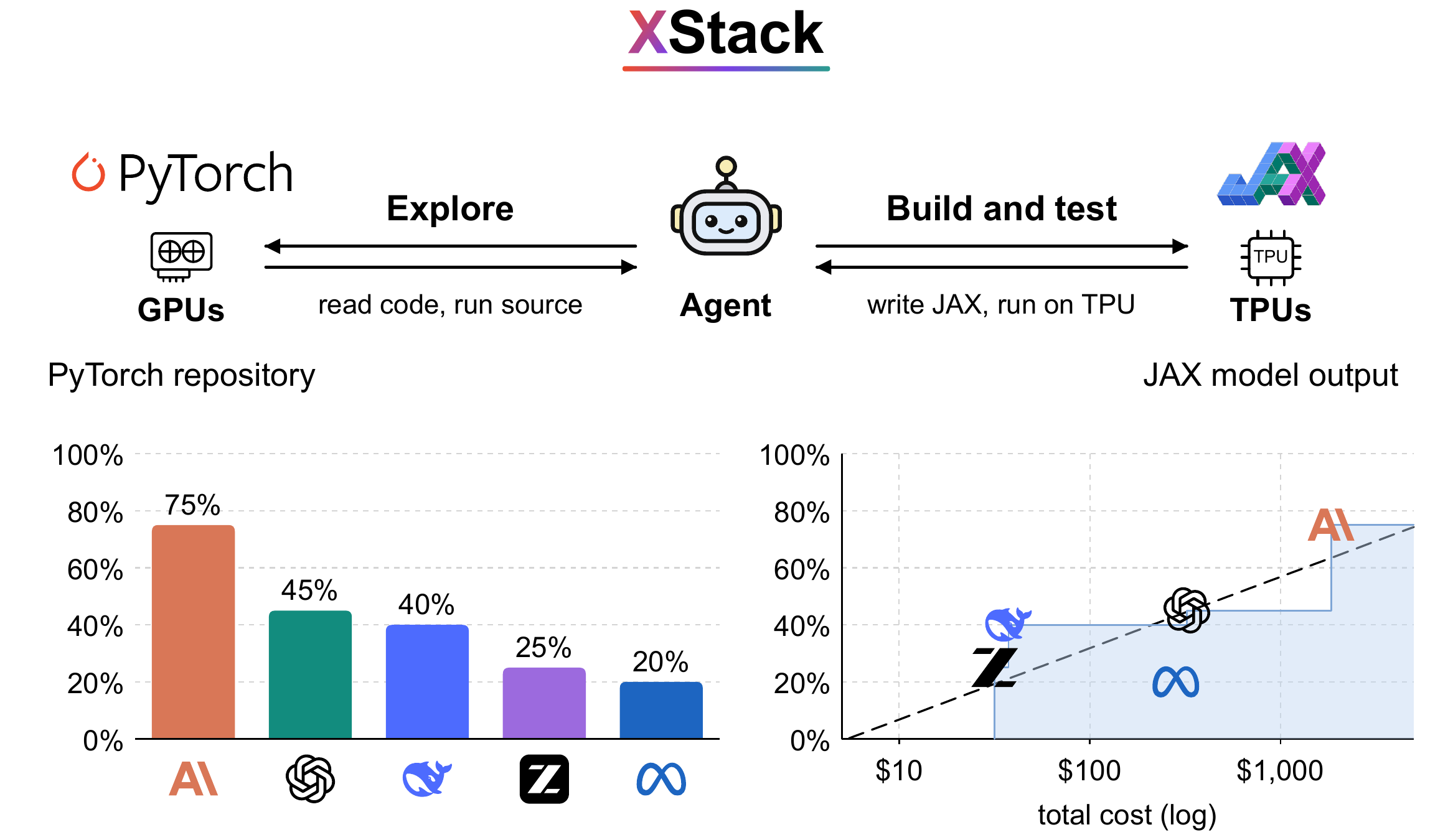}
        \caption{\textbf{Overview of \bench.} Top: in our benchmark, the agent explores a PyTorch repository built for GPUs, reads its code and runs the source, then builds and tests a JAX version on TPUs. Bottom left: pass rates of the 5 best-performing agents among the \nAgents{} we evaluate. Bottom right: the pass rates against total cost on a log scale; the dashed line is a linear fit, and the shaded steps are the Pareto frontier, the best pass rate reached at or below each cost.}
        \vspace{-0.4cm}
        \label{fig:teaser}
    \end{figure}%
}

\vspace{-1.37cm}
\linkstablestyleone
\vspace{0.1cm}
\teaserfigure{!ht}{\textwidth}
\vspace{0.4cm}
\begin{abstractblock}
\vspace{-0.7em}
\abstractcontent
\end{abstractblock}
\newpage

\section{Introduction}
\label{sec:intro}

Suppose you find a repository that does exactly what you need. Its paper reports strong results, the code is public, and the experiment has already been shown to work. There is only one problem: it was built for the wrong machine. Your lab has TPUs, but the repository assumes GPUs and PyTorch. Although, in principle, nothing about the scientific idea depends on this choice, in practice you are trapped. Using it means rewriting the model, optimizer, data pipeline, and evaluation in JAX, since that is the stack TPUs run, then making everything work under a different compiler, memory hierarchy, and distributed execution model. What looks like an implementation detail becomes a barrier to reusing the science. The hardware a team happens to have can determine which published ideas it can build on.

This is exactly the kind of barrier that coding agents should make disappear. Much published ML code is written in PyTorch for GPUs, while, for example, people also use TPUs, which are commonly programmed through JAX\footnote{JAX is Google's Python library for array computation; it compiles NumPy-style code with the XLA compiler to run on TPUs and GPUs \citep{bradbury2018jax}.}. By the end of 2025, TPUs made up nearly a fifth (18.2\%) of the world's AI compute: 3.81 million of 20.93 million H100-equivalents \citep{epoch2026chipowners}. Yet only 4.2\% of the framework-based code released with ML papers in 2024 used JAX \citep{pwc2025frameworks}. This creates a growing need to move working systems across software and hardware stacks, a real-world task that researchers care about because it unlocks access to different hardware. Doing so means crossing not only code-level framework boundaries, but also machine-level differences. This is exactly the problem we want agents to solve; we want to bring pure science back to researchers, and provide a measurement for this progress.

\begin{table}[H]
\begin{center}
\setlength{\tabcolsep}{6pt}
\begin{tabular}{ll>{\raggedright\arraybackslash}p{5.2cm}}
\toprule
 & \textbf{PyTorch on GPUs} & \textbf{JAX on TPUs} \\
\midrule
chips made by & NVIDIA and others & Google \\
where to get them & most clusters and clouds & Google Cloud or buyers' own sites \\
how the code runs & one step at a time & compiled by XLA before execution \\
how a model is written & objects that hold their weights & functions that take the weights \\
\bottomrule
\end{tabular}
\end{center}
\caption{\textbf{PyTorch on GPUs versus JAX on TPUs.} A migration changes the hardware as well as the code, so the agent must do more than translate code \citep{bradbury2018jax,paszke2019pytorch}.}
\label{tab:stacks}
\end{table}
\vspace{-0.2cm}
\vspace{-0.3cm}
We ask the following question directly:
\vspace{-0.3cm}
\begin{center}
\textit{Can an agent migrate a working ML repository to a new framework and new hardware correctly?}
\end{center}
\vspace{-0.3cm}

\paragraph{Existing benchmarks do not quite test this.} Issue-resolution benchmarks \citep{jimenez2024swebench,zan2025multiswebench} start from broken code and ask agents to fix it, while conversion benchmarks \citep{wang2024repotransbench,song2026t2j,hong2026swerefactorbench} compare translated code with the original on the same machine. In our case, the repository already works on the original machines. The challenge is to preserve that behavior while moving the entire system onto hardware the original code does not support. We introduce \textbf{XStack} (XStack stands for Cross-Stack Benchmark) to test this directly. Each task gives an agent a working PyTorch repository and a clean TPU machine, then asks it to rebuild the code on TPU with JAX and make an original experiment run end to end and reach its published result.

The answer to our question above is currently no: agents are not fully reliable for this task. Our results show that this barrier is beginning to break, but has not disappeared. The strongest agent successfully migrates 15 of 20 repositories. Agents can therefore already carry some ML systems across framework and hardware boundaries at repository scale. Yet this capability is far from reliable across agents and repositories. Agents are beginning to loosen the link between scientific code and the stack it was written for, but we are still not at a point where researchers can simply run published work wherever they choose.

A broader pattern emerges from the failures: agents can write a great deal of code, but they are still much less reliable at verifying that what they built is faithful, complete, and actually works in the real environment. This shows up across models as missed evaluations, fidelity gaps, crashes, and even attempts to work around the task when they get stuck.

Overall, our contributions are: 

\begin{itemize}
\setlength{\itemsep}{0pt}\setlength{\parskip}{0pt}
\item We introduce \textbf{\bench}, a benchmark of 20 real repositories for testing whether agents can migrate working ML systems across frameworks and hardware.
\item We evaluate seven coding agents on real TPUs and release the grading criteria and full agent trajectories, so progress on this capability is measurable.
\item We find that agents can sometimes complete these migrations, but still struggle to reliably verify that what they built is faithful, complete, and actually works end to end. We also observe that pass rate grows linearly with the log of an agent's cost (Figure~\ref{fig:cost}).
\end{itemize}

\section{The \bench Benchmark}
\label{sec:bench}

\begin{figure}[t]
\begin{center}
\includegraphics[width=.82\linewidth]{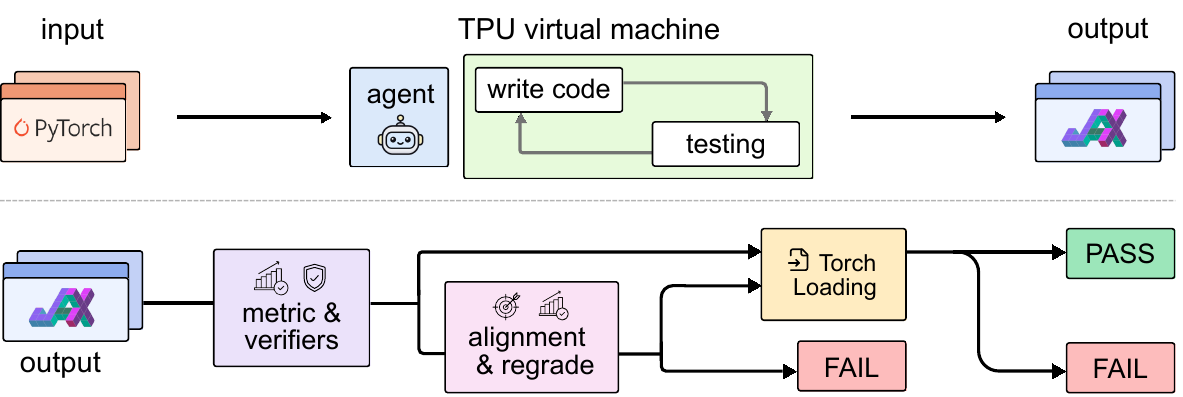}
\end{center}
\caption{\textbf{Pipeline of \bench.} Top: the agent migrates a PyTorch repository to JAX on a bare TPU. Bottom: we rerun the output on a fresh TPU and check its number against the source's result and with two model judges. If they fail, the alignment step (phase 2, pink) gives the agent the main experiment's recipe, lets it change only the entry script, and regrades the result. A pass also needs the weights to load into the original PyTorch model.}
\label{fig:pipeline}
\end{figure}

Each task in \bench is a PyTorch repository that already works and reports
a result (Figure~\ref{fig:pipeline}). The agent migrates it to JAX on a bare
TPU machine, and we run the migrated code again on a fresh machine to check
that it meets our grading criteria. For fairness, grading has
two phases: if the first attempt fails, a second phase tells the agent
which experiment we grade, so a correct migration is not failed only for
running a different experiment (Section~\ref{sec:twophase}). The numerical
target stays hidden, and only the entry script may change; all other code
and setup remain frozen. Section~\ref{sec:task} defines the
task, Section~\ref{sec:grading} explains how we grade it, and
Section~\ref{sec:repos} describes our repositories and agents.

\subsection{Task Formulation}
\label{sec:task}

\textbf{Input.} The agent receives the repository, pinned to one git
commit so that every agent sees the same code, and a TPU machine. One prompt
template, the same for every repository, asks it to migrate the entire repository to JAX, with Flax for the
model and Optax for the optimizer
\citep{bradbury2018jax,heek2020flax,deepmind2020jax} (Table~\ref{tab:stacks}), and to run it the way
the source runs its experiments, at full scale. For four repositories whose source offers several configurations, one or two added lines name the graded configuration or a flag the grader uses, so that every agent knows what it has to do without learning the grading target (Appendix~\ref{app:prompts}).

\textbf{Machine setup.} The machine has no machine learning software; setting it up is part of the task: the agent installs every package, fetches
the data and decides how to use its hosts. It works from a sandbox
on our cluster, reaches the machine over SSH and has no time limit while it
produces output (Appendix~\ref{app:isolation}).

\textbf{Output.} The agent delivers the migrated repository with three
scripts. \file{setup\_env.sh} prepares a clean machine, the entry script
\file{run\_train.py} trains, evaluates and writes metrics, and
\file{export\_torch.py} saves the weights under the source's parameter
names.\label{sec:info} In phase 1 the agent is not
told which number we grade, what value to reach or whether it passed
(Table~\ref{tab:info}), so it has to migrate the whole repository, not one
experiment.

\subsection{Grading a Migration}
\label{sec:grading}
\label{sec:isolation}

\textbf{Reaching the grading target.} We copy the migrated repository, without its git history, to a freshly wiped machine and run \file{setup\_env.sh} and \file{run\_train.py} once (the purple stage of Figure~\ref{fig:pipeline}). A script, the grader, compares the graded number with the grading target, usually the number the source published, with a tolerance (Appendix~\ref{app:fleet}). Each grading target follows the source's published result, which we authenticated against the pinned source; for \nBarAnchor{} repositories we set it from our own runs of the unmodified source, and for minGPT as a rule (Section~\ref{sec:repos}). A faithful migration reaches the grading target by reproducing the repository, without knowing the target. A run that writes no metrics within ten times its expected run time fails.

\textbf{Judges.} A number can also reach the grading target for the wrong reason, such as a smaller variant or another run's files. Two model judges, both GPT-5.6-sol, therefore review every result that reaches the grading target (the verifiers of Figure~\ref{fig:pipeline}). The correctness judge checks that the number comes from this run's training of the main experiment, and the cheat judge checks that the agent did not reach it by cheating, for example by clamping the metric or using another run's files. A result passes only if it reaches the grading target, both judges accept it, and its weights load into the source model.

\begin{figure}[t]
\begin{center}
\input{figures/fig2_fleet_links.tex}
\end{center}
\caption{\textbf{Repositories of \bench.} Rows group the \nSlate{} repositories into three task families. The number beside each name is the expected duration, in hours, of one graded run, which trains the model at the source's full scale.}
\label{fig:fleet}
\end{figure}

\begin{table}[t]
\centering
\newcommand{\yes}{\textcolor{passgreen}{$\checkmark$}}
\newcommand{\no}{\textcolor{failred}{$\times$}}
\setlength{\tabcolsep}{5pt}
\begin{tabular}{lccc}
\toprule
 & \multicolumn{2}{c}{agent} & \\
\cmidrule(lr){2-3}
 & phase 1 & phase 2 & grader \\
\midrule
source repository & \yes & \yes & \yes \\
TPU machine & \yes & \yes & fresh \\
task prompt & \yes & \yes & \no \\
experiment recipe & \no & \yes & \yes \\
grading target & \no & \no & \yes \\
grading code & \no & \no & \yes \\
outcome & \no & \no & \yes \\
\bottomrule
\end{tabular}
\caption{\textbf{What the agent and the grader see.} The agent never sees the grading target, so it has to migrate the whole repository; phase 2 adds the main experiment's recipe to ensure grading fairness.}
\label{tab:info}
\vspace{-0.3cm}
\end{table}

\textbf{Two phases.}\label{sec:twophase} With the criteria hidden, an agent can migrate the code correctly and still run an experiment we do not grade, such as AST's ESC-50 recipe instead of Speech Commands
\citep{warden2018speechcommands}. Scoring that as a failure would penalize a choice the task never specified. Phase 2 removes this unfairness without making the task easier: after a failed phase 1, one message (the alignment step of
Figure~\ref{fig:pipeline}) freezes everything except
the entry script and gives the recipe of the main experiment, but not the
grading target (Table~\ref{tab:info}). Changing any other file fails phase 2 as a
\emph{freeze violation}, and the judges fail a pass whose entry script
rebuilds the frozen code, a broken freeze. We report the pass rate after
phase 2 as the main result.

\vspace{4pt}
\subsection{Repositories and Agents}
\label{sec:repos}
\label{sec:setup}

\textbf{Repositories.} \bench contains \nSlate{} repositories from fourteen
domains (Figure~\ref{fig:fleet}; Table~\ref{tab:fleet} gives each grading target).
Seven are language models: minGPT \citep{karpathy2020mingpt}, nanoGPT
\citep{karpathy2022nanogpt}, makemore \citep{karpathy2022makemore},
llama2.c \citep{karpathy2023llama2c}, modded-nanoGPT
\citep{jordan2024moddednanogpt}, nanochat \citep{karpathy2025nanochat} and
x-transformers \citep{wang2020xtransformers}. The others are LoRA
\citep{hu2022lora}, SimCSE \citep{gao2021simcse}, airbench
\citep{jordan2024airbench}, AST \citep{gong2021ast}, S4 \citep{gu2022s4},
HSTU \citep{zhai2024hstu}, TD-MPC2 \citep{hansen2024tdmpc2}, TensoRF
\citep{chen2022tensorf}, nanoVLM \citep{wiedmann2025nanovlm}, deepspeech
\citep{naren2017deepspeech}, TGN \citep{rossi2020tgn}, PatchTST
\citep{nie2023patchtst} and TabR \citep{gorishniy2024tabr}. We chose
them from more than 200 PyTorch repositories, most of which we
rejected because of their published number.

\textbf{Grading targets.} For \nBarPublished{} repositories, the grading target is the number
that the README or paper reports, which we authenticated against the source at its pinned commit; deepspeech's comes from its release notes. A fair target must be one that the original code itself reaches, and our experiments confirm that each of these targets can be reached. For \nBarAnchor{} others, we set the target from our own GPU runs of the unmodified source. For example, the makemore README reports a loss of about 1.92, but eight runs of the unmodified source on our GPUs average 1.989 with a standard deviation of 0.017, so a target of 1.92 would fail the original code itself. We therefore set its target to a best test loss of at most 2.023, the mean plus two standard deviations. For minGPT, whose task trains a small GPT to add two-digit numbers, we require at least 495 of the 500 held-out sums to be correct: a trained model is expected to solve all 500, and the margin allows five misses for seed and numerical noise.

\begin{figure}[t]
\begin{center}
\includegraphics[width=\linewidth]{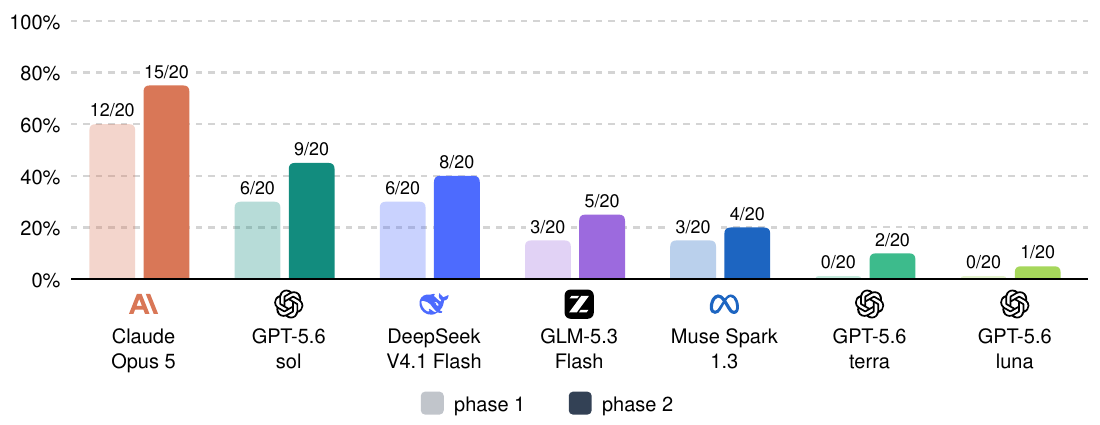}
\end{center}
\caption{\textbf{Pass rate per agent.} Light bars: after phase 1; dark bars: after phase 2. Phase 2 adds a few passes for every agent: once aligned with the main experiment, each agent passes more repositories.}
\label{fig:accuracy}
\end{figure}

\begin{table}[t]
\begin{center}
\setlength{\tabcolsep}{3pt}
\begin{tabular}{llcl}
\toprule
agent & harness & reasoning effort & list price \\
\midrule
\input{figures/table_agents.tex}
\end{tabular}
\end{center}
\caption{\textbf{Harness, reasoning effort and price per agent.} All agents run at high reasoning effort with the same settings, and every model is called through its developer's official API. Prices are the official list prices on \svPriceDate, USD per million tokens (input / cached input / output).}
\label{tab:agents}
\end{table}

\textbf{Agents and cells.} We evaluate \nAgents{} agents from \nLabs{} labs
(Table~\ref{tab:agents}): \mbox{GPT-5.6-sol}, \mbox{GPT-5.6-terra} and
\mbox{GPT-5.6-luna} \citep{openai2026gpt56card}, Claude Opus 5
\citep{anthropic2026opus5card}, GLM-5.3-Flash
\citep{zai2026glm53flash,glm5team2026glm5}, DeepSeek V4.1 Flash
\citep{deepseekai2026v41flash} and Muse Spark 1.3
\citep{meta2026musespark13}, all at high reasoning effort. We call a pair of an agent and a repository a \emph{cell}.
Each cell runs on a TPU v6e, v5p or v4 machine
\citep{jouppi2023tpuv4,googlecloud2026tpuv5p,googlecloud2026tpuv6e}, and we
ensure every repository has adequate compute
(Appendix~\ref{app:candidates}; Figure~\ref{fig:cand-code-and-compute}).

\section{Results Analysis}
\label{sec:results}

\debugpar

\subsection{Main Results}
\label{sec:single}

\textbf{Phase 1.} In phase 1, the agents pass
\svPhaseOneAll{} cells (\svPhaseOneAllPct; Figure~\ref{fig:accuracy}). The
grader passed \svGraderBlind{} phase~1 results, and the judges rejected
\svJudgeFailedBlind{} of these. Claude Opus 5 leads by a wide margin with
\svPhaseOneClaude. DeepSeek V4.1 Flash follows with \svPhaseOneDeepseek, and
among the GPT-5.6 tiers only sol passes any repository (\svPhaseOneSol). Migrating a whole repository is thus still hard for current agents: most attempts fail, even though every agent has the full source and its own TPU, on which it can test extensively.

\textbf{Phase 2} runs on \svPhaseTwoRan{} cells that failed phase 1. The grader passed \svGraderRescue{} of them, and the judges let
\svPhaseTwoRescued{} of these pass. These are the cases the phase was designed
for, such as a fixed number of steps where the source trains until a stop
rule, a training statistic reported where the source runs an evaluation, or another variant of the source's recipe.
The judges rejected \svJudgeFailedRescue{} phase 2 passes,
\svJudgeBroken{} of them because the entry script rebuilt the frozen code,
and the remaining phase 2 failures lie in the frozen code
(Section~\ref{sec:twophase}). After phase 2, \svHeadlinePasses{} cells pass the grader and the judges. The weights of \svAuditFlagged{} of them do not load under the source's parameter names, so the agents pass \svFullAll{} cells (\svFullAllPct).
Appendix~\ref{app:cells} lists every cell's result.

Phase 2 aligns the agent with the recipe of the experiment
we grade, so a correct migration that first ran a different experiment is
not failed for that choice. Every agent gains at least one pass from it. It
also helps ensure the remaining failures lie in the migrated code, not
in an under-specified task (Section~\ref{sec:twophase}). Running all \svCells{} cells took \svTpuChipHours{} TPU chip-hours
(Figure~\ref{fig:cand-code-and-compute}), and the agents' trajectories sum to
\svTotalTokensB{} billion tokens.

\subsection{Judges, Cost and Behavior}

\label{sec:audit}

\begin{figure}[t]
\begin{center}
\includegraphics[width=\linewidth]{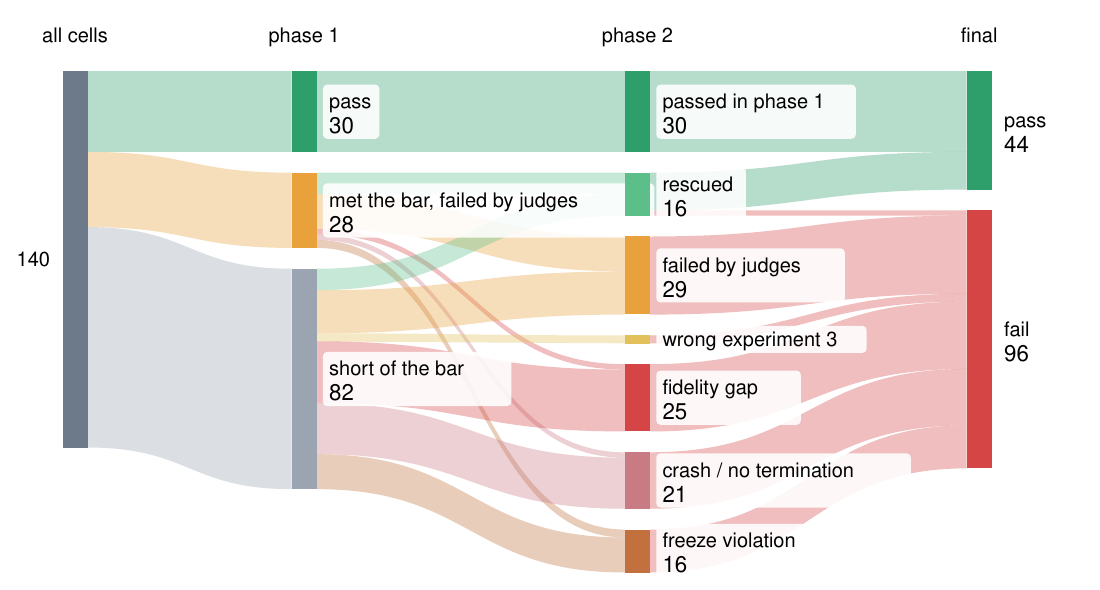}
\end{center}

\caption{\textbf{Outcomes by phase.} From left to right, the columns show all \svCells{} cells, their phase~1 outcome, their phase~2 outcome and the final result. Of the \svTaxFailsOne{} cells that failed phase~1, phase~2 rescues \svPhaseTwoRescued{}; the rest fail the judges, fall short of the grading target, crash or break the freeze (Section~\ref{sec:twophase}). Two rescued cells fail the weight-loading check.}
\label{fig:funnel}
\end{figure}

\paragraph{Effect of the judges.}
The judges are needed, because they catch results that reach the
grading target for the wrong reason, which the metric alone cannot detect.
Across both phases, they rejected \svJudgeFailed{} of the \svGraderPasses{}
results that the grader passed (Figure~\ref{fig:funnel};
Table~\ref{tab:gates}). We checked many rejections by hand and found
them sound; the judge caught a llama2.c migration that wrote the published
number into its metrics beside its own and a minGPT migration evaluated on
its training data.

\textbf{What the judges find.} Apart from broken freezes,
the judges catch two kinds of false positives. The grader reads whatever
number the run reports, which may not be the recipe's own, such as a
separate validation pass in place of the recipe's averaged loss; the
correctness judge finds the recipe's number in the run's logs and checks
it, so every pass is a true positive. The cheat judge catches cheating, such as training on another task's machine or reading
another session's files.

\label{sec:cost}

\begin{figure}[t]
\begin{center}
\includegraphics[width=\linewidth]{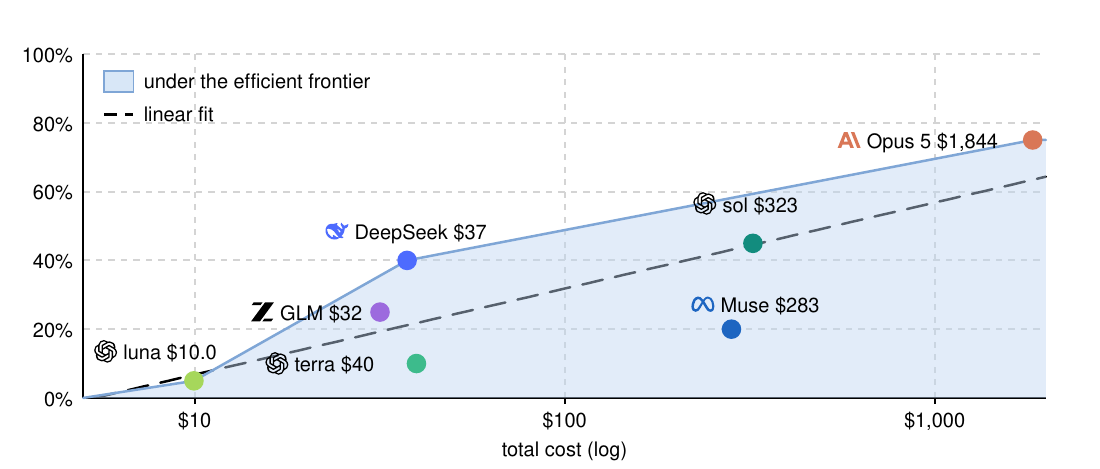}
\end{center}
\caption{\textbf{Accuracy against cost.} Each dot is one agent's final pass rate against its cost at the list prices of \svPriceDate{} (Appendix~\ref{app:cost}). The dashed line is a least-squares fit of pass rate on log cost. The blue line is the efficient frontier, the upper convex hull of the agents with cost on a log scale. Pass rate rises by about \svCostFitSlope{} percentage points with every $10\times$ in cost.}
\vspace{-0.3cm}
\label{fig:cost}
\end{figure}

\textbf{Accuracy versus cost.} Pass rate grows linearly with the logarithm of cost
(Figure~\ref{fig:cost}). We fit pass rate on log cost because the agents'
costs for the \nSlate{} repositories span from \svTotalCostLuna{} to \svTotalCostClaude. Every $10\times$ in cost adds about
\svCostFitSlope{} percentage points, and the line explains most of the
difference between agents ($R^2 = \svCostFitRTwo$, $r = \svCostFitR$,
$p = \svCostFitP$, $n = \svCostFitN$ agents). The trend is statistically significant at the 5\% level. The upper edge of the shaded area, the efficient frontier,
runs through three agents: \mbox{GPT-5.6-luna}, the cheapest; DeepSeek V4.1
Flash, the cheapest per pass; and Claude Opus 5, the most accurate. Every
other agent lies below it; Muse Spark 1.3, for example, costs more than seven times as
much as DeepSeek V4.1 Flash and passes fewer repositories.

\begin{table}[H]

\begin{center}
\setlength{\tabcolsep}{3pt}
\input{figures/table_checks_cost.tex}
\end{center}
\caption{\textbf{Checks and cost per agent.} \textbf{Left}: share of repositories passing each check in turn (grading target, correctness and cheat judges, weight loading). \textbf{Right}: repositories passed, mean tokens per cell (input, cached input and output over all turns, as billed), list-price dollars per cell and per pass, and mean agent hours per cell.}\label{tab:gates}\label{tab:profile}
\end{table}

\paragraph{Behavior difference between agents.}
Reading the trajectories, we found that Claude~Opus~5 differs from the
other agents in how it checks its work: before handing in, it runs the
source's experiment to the end and compares its number with the published
one. This checking takes time and tokens, and Opus is both the
most expensive agent and the one that passes the most
(Table~\ref{tab:profile}). On nanochat, \mbox{GPT-5.6-sol} estimated the full run at a few hours
and handed in without running it. Muse Spark 1.3 does
run the full experiment, but on LoRA it handed in a BLEU score far below
the paper's and called its migration ``fully validated''.
We find that diligent testing and experimentation, together
with careful comparison against the original code, are what a migration
needs to succeed.

\section{Why Do Agents Fail?}
\label{sec:analysis}

\subsection{Failure Taxonomy}
\label{sec:taxonomy}

\begin{figure}[t]
\begin{center}
\includegraphics[width=\linewidth]{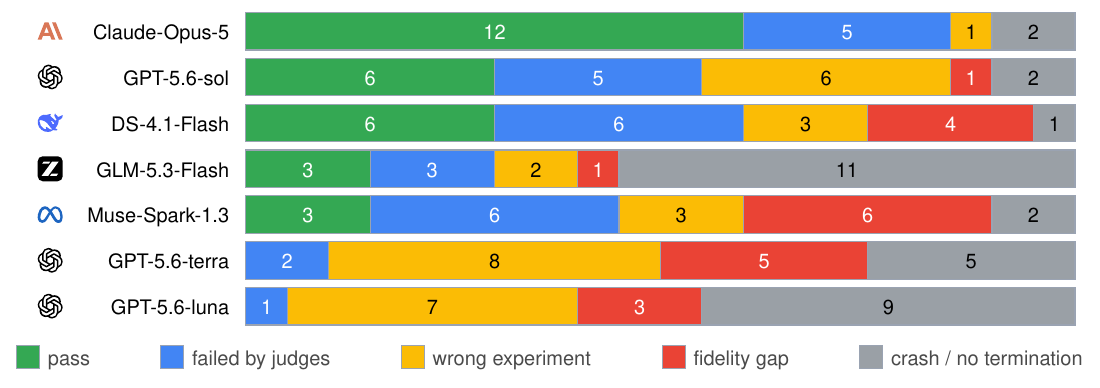}
\end{center}
\caption{\textbf{Failure breakdown after phase 1.} Each bar splits an agent's \nSlate{} repositories by phase 1 outcome: pass; failed by the judges; wrong experiment (a different pipeline or scale, or the headline evaluation never run); fidelity gap (right pipeline, misses the grading target); crash or no termination. The last two, \svTaxCodePctOne{} of failures, are faults in the migrated code, which phase 2 cannot repair.}
\label{fig:taxonomy}
\end{figure}

A third judge, the failure judge, assigns every graded run one class
from a fixed taxonomy (Appendix~\ref{app:taxonomy}), based on its
metrics, its log and the diff of the migrated code. The
failure judge only reads the run; it cannot determine whether
the run passes or fails. Figure~\ref{fig:taxonomy} shows the failure judge's classification of phase~1 runs.

\textbf{The code fails, not the choice of experiment.} We group the classes
by what could fix them. A wrong or incomplete experiment (a different
pipeline, the right one at a different scale, or a run that never reaches
the main evaluation) is a problem with what the entry script runs, and phase
2 can fix it. A fidelity gap, a crash or a run that never ends is a problem
in the migrated code. In phase 1, \svTaxWrongPctOne{} of the
failures are wrong or incomplete experiments and \svTaxCodePctOne{} are code
failures; the judges reject the rest. Phase 2 removes the first group.
The cells it does not rescue still fail in the code, or they break the
freeze by rebuilding the migrated code inside the entry script
(Figure~\ref{fig:taxonomytwo}). Aligning the agent with the grading recipe does not make a cell pass if its migration is wrong in the first place.

\textbf{Agents fail differently.}
DeepSeek V4.1 Flash rarely crashes (\svCrashDeepseek{} cell),
and \svGateBarDeepseek{} of its cells reach the grading target, but the
judges reject half of these, mostly for cheating. Terra's failures are spread over wrong experiments, fidelity gaps and crashes, and luna's over almost every class (Figure~\ref{fig:taxonomytwo}).

\textbf{Translation alone is not the hard part.} Because JAX
also runs on GPUs, we can also test translation between PyTorch and JAX
alone, on the same hardware. In the earlier development of our
benchmark, we asked for a whole JAX repository and graded it on GPU. There
the migration was nearly solved: GPT-5.5
at high reasoning effort migrated 12 PyTorch repositories to JAX and passed
99\% of the checks, 116 of 117, with 11 of the 12 repositories passed in full
(Appendix~\ref{app:earlier}, Table~\ref{tab:gpu-pilot}). The checks are
strict: forward and gradient parity with PyTorch, checkpoints that load in
both frameworks, decoding that matches token for token, and one miss fails
the repository.

On TPUs, even after phase~2, the agents' average pass rate
is only \svFullAllPct, and about half the failures are code that
crashes or falls short on a machine the agent set up. The comparison
suggests the difficulty lies not in translation alone but in migrating
across both framework and hardware. Testing this ability therefore requires
migrating across hardware, which is exactly what \bench does.

\subsection{Behavior and Integrity}
\label{sec:behaviour}
\label{sec:integrity}

Agents that pass follow the source's documentation, running
the README's default recipe, reporting its numbers under its names, and
getting results close to the source's.
Agents that get stuck explore the machine instead, reading other cells'
leftovers or using our credentials. We found these leaks in our initial
experiments, hardened the sandbox, and reran the three affected cells cleanly.
Two incidents that aimed at the grading
target, a clamped metric and a copied checkpoint, also date from building
the benchmark and affect neither the current benchmark nor its results. In the final benchmark, cheating incidents instead use what the source offers, such as a sibling variant or the published number, and our checks and the cheat judge catch them.

\section{Related Work}
\label{sec:related}

\paragraph{Code migration.}
Code translation has grown from single functions and small programs
\citep{roziere2020transcoder,pan2024lost} to whole repositories
\citep{ibrahimzada2025alphatrans,wang2024repotransbench,li2026repomodbench}, although some
repository benchmarks still grade one function in context
\citep{ou2025rustrepotrans} or fill in a skeleton
\citep{zhang2025skeleton,khatry2025crustbench}. Migration benchmarks upgrade
a language version \citep{liu2025migrationbench} or move a repository onto another software stack
\citep{hong2026swerefactorbench}.

\paragraph{Framework conversion.}
For machine learning code, framework
conversion has been studied for single snippets \citep{phan2026t2j}, short programs
\citep{yan2023codetransocean} and training codebases \citep{song2026t2j}, or avoided
with a portability layer \citep{lenton2021ivy}.
T2J-Bench \citep{song2026t2j} is the closest: whole PyTorch training
codebases to JAX under a hidden evaluator, but on one GPU, the framework-only
setting our GPU experiments found nearly solved (Section~\ref{sec:taxonomy}). Kernel benchmarks write kernels for one accelerator
\citep{ouyang2025kernelbench,tschand2026jaxbench,wang2026maxkernel}, one operator
at a time against a reference that already runs on it.
Models do worse on hardware rare in their training data
\citep{wen2025multikernelbench}. In all of these the accelerator stays
the same or the unit is one kernel. \bench moves a whole repository to a new
framework and hardware and forbids PyTorch at runtime, so no agent passes
by doing nothing.

\paragraph{Agent benchmarks.}
Since SWE-bench \citep{jimenez2024swebench}, issue-resolution benchmarks have
graded a change with tests that fail before it and pass after it. Later
benchmarks ask for a codebase built from scratch
\citep{zhao2025commit0,yang2026programbench}.
A working
repository, however, has no failing test in its own environment. Most also give
the agent a ready environment; SetupBench \citep{arora2025setupbench} and
DeployBench \citep{wang2026deploybench} test setup alone, but stop once
the code runs. Closer to our setting are benchmarks graded against research
results. Agents reproduce a paper's results from the authors' code
\citep{siegel2024corebench}, replicate papers against rubrics or reference
code \citep{starace2025paperbench,seo2025paper2code}, or maximize a score on
machine learning tasks \citep{chan2025mlebench,wijk2025rebench}.
Our grading target, in contrast, is the source repository's own result
(Section~\ref{sec:repos}), which a faithful migration will naturally reach.

\paragraph{Evaluation integrity.}
Agents exploit underspecified objectives
\citep{krakovna2020specification,skalse2022defining} and game checks they
can see or tamper with \citep{zhu2025abc,wang2026benchjack}; benchmarks
built from public code risk memorization
\citep{jain2025livecodebench,liang2025swebenchillusion}. In \mbox{\bench}, the grading
target is hidden: the agent is never told it or whether it passed. For
fairness, after a failed first attempt a second phase aligns the agent with
the graded experiment while its code, except the entry script, stays frozen. We rerun the code on a fresh machine, so
remembering an answer is not enough, and model judges review every result
that reaches the target (Section~\ref{sec:isolation}).

\section{Conclusion}
\label{sec:conclusion}

We introduce \textbf{\bench}, a benchmark that evaluates agents' ability to migrate working ML repositories from PyTorch on GPUs to JAX on TPUs. Across \nAgents{} agents from \nLabs{} labs, the average pass rate is
\svFullAllPct, and pass rate grows linearly with the
logarithm of cost. Our analysis shows that verification separates success
from failure: agents that run the source's experiment to the end and compare
their result with the original tend to pass, and agents that hand in
untested code tend to fail. We encourage other benchmarks
with hidden criteria to adopt a similar frozen second phase that aligns the agent with what is graded, so that they stay fair to agents. Although
current agents are not yet reliable at migrating across both frameworks and hardware, we hope they will one day handle this task reliably and unlock access to different hardware for researchers. In the meantime, \bench will help measure progress toward this hardware freedom.

\section*{Acknowledgements}
This work was supported by Princeton Research Computing resources, including the Della high-performance computing cluster. We thank Princeton Language and Intelligence (PLI) for their support of this research, as well as Google's TPU Research Cloud (TRC) program for providing TPU compute.

\bibliographystyle{plainnat}
\bibliography{xstack}

\clearpage

\appendix
\makeatletter
\renewcommand\thetable{\thesection\@arabic\c@table}
\renewcommand\thefigure{\thesection\@arabic\c@figure}
\renewcommand\thelstlisting{\thesection\@arabic\c@lstlisting}
\@addtoreset{figure}{section}
\@addtoreset{table}{section}
\@addtoreset{lstlisting}{section}
\makeatother
\setcounter{figure}{0}
\setcounter{table}{0}

{\huge\bfseries Appendix}\par
\vspace{1cm}
\noindent The appendix holds the repositories, headline metrics and grading targets
(\ref{app:fleet}); the agent's inputs, verbatim (\ref{app:prompts}); the
sandboxing (\ref{app:isolation}); the failure taxonomy definitions
(\ref{app:taxonomy}); per-cell results (\ref{app:cells}); cost (\ref{app:cost});
the GPU-only experiment (\ref{app:earlier}); and code size and compute
(\ref{app:candidates}).

\begingroup
\hypersetup{linkcolor=black}
\etoctoccontentsline{part}{}
\etocstandardlines
\etocsettocstyle{}{}
\etocsetnexttocdepth{subsection}
\localtableofcontents
\endgroup

\section{Repositories, Headline Metrics and Grading Targets}
\label{app:fleet}

\paragraph{Sourcing.} Hard requirements: a training-centric pipeline that is
the repository's point; a readable, vendorable tree at a pinnable commit under
a permissive license read from the LICENSE file, not the GitHub label; a
headline metric that the repository's own code computes, for a configuration the tree actually runs, with a published value; anonymously acquirable data; a full-scale run of about ten hours or less on one TPU host.

\paragraph{Notation.} Table~\ref{tab:fleet} is generated from the criteria
files the grader reads. \emph{Grading target} uses the comparator's notation: $\le$ and $\ge$
are one-sided, $[x, y]$ is a band whose upper bound is the metric's ceiling
or, on a loss, a leakage floor. \emph{Origin}: \emph{readme} or \emph{paper}, the
source's own published number for the configuration in the tree;
\emph{anchor}, a pre-registered band around our runs of the unmodified
source; \emph{rule}, a floor we set for a repository that publishes no number;
\emph{released-checkpoint}, the number the source's release notes report for
its pre-trained checkpoint.
The table in the
release quotes each statement verbatim with its line pointer.

{\setlength{\tabcolsep}{2.5pt}\setlength{\LTcapwidth}{\linewidth}
\begin{longtable}{>{\raggedright\arraybackslash}p{2.25cm}>{\raggedright\arraybackslash}p{2.7cm}>{\raggedright\arraybackslash}p{4.28cm}p{2.32cm}>{\raggedright\arraybackslash}p{1.7cm}}
\toprule
repository & domain & headline metric & grading target & origin \\
\midrule
\endfirsthead
\toprule
repository & domain & headline metric & grading target & origin \\
\midrule
\endhead
\bottomrule
\endfoot
\bottomrule
\noalign{\vskip 8pt}
\caption{The \nSlate{} repositories by domain family, with the graded quantity and its grading target.}\label{tab:fleet}
\endlastfoot
\input{figures/table_fleet.tex}
\end{longtable}
}

\section{Agent Inputs, Verbatim}
\label{app:prompts}

A cell's input is the rendered prompt, the access document with live
addresses, and the vendored repository. Below, typography is
normalized and lines re-wrapped to fit the page; the words are verbatim. The recipe
paragraph of every repository, the three judge prompts and every judge answer
are released with the benchmark; the prompts are in the code at
\href{https://github.com/zlab-princeton/XStack}{github.com/zlab-princeton/XStack} under \file{prompts/}.

\subsection{Prompt Template}

\begin{tcblisting}{promptstyle, title={Phase 1 Prompt Template}}
# PyTorch -> JAX on TPU: whole-repo transfer

Port the fixed-commit PyTorch repository under `source_torch/` entirely to a working JAX/Flax/Optax repository, and make it work at real scale on the TPU machine described in `TPU_ACCESS.md` -- a clean machine: set it up yourself. You have full access to that machine to develop, run, and test your port. The deliverable is the working repository itself: produce a faithful port and run it the way the source does.

Deliverables at this workspace root:

- `setup_env.sh` -- prepares a worker of the TPU machine from clean state, including any data the repository's pipeline needs, obtained and prepared the way the source repository does. It is sourced before the entry point runs.
- `run_train.py --seed S --out DIR --metrics PATH` -- runs the repository's training pipelines the way the repository runs them, at their own scale, and writes the numbers the repository itself reports to the metrics JSON. It checkpoints to `--out` as it trains and resumes from an existing checkpoint in `--out`.
- `export_torch.py --ckpt DIR --out PATH` -- writes the trained weights as an `.npz` of numpy arrays keyed by the source model's own parameter names (exact names and shapes).
{{REPO_CONTRACT}}
Your submission is collected from this workspace and executed on a fresh machine of the same type. It is judged on behavior that is fundamental to the source repository, not all of which is spelled out here.

Rules:

- Do not import or call PyTorch anywhere in your submission at runtime.
- Data, library, and package downloads are allowed; do not fetch, clone, or copy model or porting code from the internet.

## Goal

Keep working until your submission passes grading.
\end{tcblisting}

The slot \file{\{\{REPO\_CONTRACT\}\}} renders empty for sixteen of the \nSlate{} repositories.
For nanoGPT it renders two lines: a \file{--max-iters N} knob the entry must honor
(the grader passes 120{,}000) and the line naming GPT-2 124M on OpenWebText as
the graded configuration. For HSTU it renders one configuration line (HSTU-large on
MovieLens-20M). For TD-MPC2 it renders \file{--task NAME} and
\file{--max-steps N} knobs (the grader passes walker-run and 500{,}000; the
prompt names neither). For nanochat, the same \file{--max-iters} knob, dormant.

\subsection{Access Document}

The multi-host document; a single-host twin exists for one-host machines:

\begin{tcblisting}{promptstyle, title={Access Document (multi-host)}}
# TPU access

A multi-host TPU machine is allocated to this task for its duration. It is a standing machine that must stay up (rules at the end).

- Google Cloud v5p slice with 8 worker hosts, 4 TPU chips each (32 chips total). A JAX program launched simultaneously on ALL workers sees all 32 devices as one machine.
- Each worker: Ubuntu 22.04, remote user `<user>`, stock Python 3.10, no ML stack preinstalled, its own ~100 GB local disk (set up every worker; large datasets and repeated checkpoints can exhaust it).
- TPU name `<name>`, zone `<zone>`. Worker external IPs: <list>

Run a command on worker N (this form is the reliable one):

    gcloud alpha compute tpus tpu-vm ssh <user>@<name> \
      --zone <zone> --worker=N --command 'command' --quiet \
      --ssh-flag=-i --ssh-flag=$HOME/.ssh/<key>

Copy files to worker N:

    gcloud alpha compute tpus tpu-vm scp --recurse local_path \
      <user>@<name>:remote_path --zone <zone> --worker=N --quiet \
      --scp-flag=-i --scp-flag=$HOME/.ssh/<key>

Direct `ssh -i ~/.ssh/<key> <user>@<worker-ip>` also works but times out intermittently in this zone; if you use it, set generous connect timeouts and retry. Prefer the gcloud form.

Notes:

- To use the whole machine, launch the SAME program on ALL workers at the same time (e.g. loop the command over workers with `&`); on TPU the JAX runtime discovers the other hosts automatically.
- Every worker has internet access; installing packages works on each.
- A dropped ssh session kills its foreground child: run long remote commands under `nohup` (or `tmux`) and poll their logs.
- Your final submission is collected from THIS workspace, then run on that machine (launched on all workers simultaneously).

Rules (the machine is shared infrastructure and must stay up):

- Never reboot, shut down or power off a worker, and never stop or restart system services. If the TPU looks stuck, a process of yours is still holding the devices: kill your own processes on every worker and wait for them to exit.
- All workers are yours to use. On each of them, keep your files under the `<user>` home directory, /tmp and /dev/shm, and only touch processes you started.
\end{tcblisting}

The document is re-rendered with live addresses before every turn.

\newpage
\subsection{Frozen Phase 2 Message}

After a phase 1 fail the agent receives nothing but this message: no
failure sentence and no pass or fail. The frame is one sentence followed by the repository's recipe paragraph:

\begin{tcblisting}{promptstyle, title={Frozen Phase 2 Message}}
Your submission is now frozen: from this point do not change your ported code, setup, or any other file. Edit only `run_train.py` so that one invocation does exactly the following, using your ported code as it is:

{{RECIPE}}
\end{tcblisting}

\section{Sandbox and Isolation}
\label{app:isolation}

The agent runs on our cluster inside a sandbox that hides the benchmark
repository, our reference migrations, the grading criteria, the results archive and the session files of every other cell. The sandbox also turns off the agent's memory across sessions. The TPU
machine itself stores no results: every graded turn is archived elsewhere, and the hosts are wiped before every cell.
An agent turn has no time limit while output flows: thirty minutes of
silence after activity ends the turn as a hang, and thirty minutes with no
output at all counts as a provider outage and the turn is retried.

\section{Failure Taxonomy Definitions}
\label{app:taxonomy}

The failure judge assigns one of the classes below.

\begin{center}
\begin{tabular}{lp{9.4cm}}
\toprule
class & meaning \\
\midrule
pass & grading target reached (a pass in the figures also needs the judges) \\
scope miss & a different pipeline than the headline one, or only part of the repository (nanoGPT's Shakespeare quick-start instead of GPT-2 on OpenWebText; minGPT's chargpt instead of the adder; HSTU's ML-1M example) \\
incomplete transfer & the right pipeline trained, the headline evaluation never run (nanochat without the CORE suite; LoRA without beam decoding and BLEU) \\
fidelity gap & the right pipeline and evaluation, the model short of the grading target \\
scale deviation & the recipe run at a different scale than the source (batch, steps, model size, data ratio) \\
runtime crash & the graded run exited non-zero or hung \\
no termination & the migration has no stop rule; the run reached the cap and wrote no metrics \\
no TPU & the graded run never acquired the TPU (CPU fallback) \\
wrong fix & a later turn that changed something other than the defect \\
unchanged tree & the agent returned without editing the deliverable \\
grader artifact & the grade was the harness's error, not the migration's; the class describes the grader \\
fabricated result & the headline number was not produced by running the pipeline (copied, hard-coded, or evaluated on downloaded finished weights) \\
freeze violation & a phase 2 turn changed something other than the entry file \\
infra & machine or harness outage; the turn was not consumed \\
\bottomrule
\end{tabular}
\end{center}

A turn may carry a secondary class. Rows are added when the grade lands and
never change it.

\begin{figure}[h]
\begin{center}
\includegraphics[width=\linewidth]{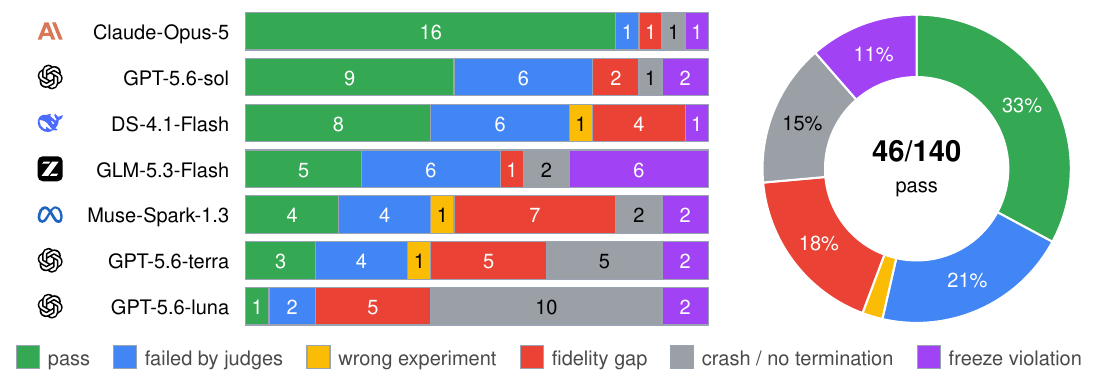}
\end{center}
\caption{\textbf{Failure breakdown after phase 2.} Outcome of each cell after phase 2, grouped as in
Figure~\ref{fig:taxonomy} plus freeze violations: runs that
changed more than the entry script or rebuilt the migration in it. The ring at the right sums the bars over the agents. Passing in either phase counts as a pass. A cell without a phase 2, or whose
phase 2 left the entry script unchanged, keeps its phase 1 outcome.}
\label{fig:taxonomytwo}
\end{figure}

Figure~\ref{fig:taxonomytwo} repeats the phase-1 breakdown of
Figure~\ref{fig:taxonomy} after phase 2. The wrong-experiment group
disappears, because phase 2 states the recipe: a cell that ran the wrong
pipeline in phase 1 either passes now or fails in its migrated code. What
remains is the two code groups, fidelity gaps and crashes, which no entry
script can fix, the judges' rejections, and the \svTaxFreezeTwo{} cells
that broke the freeze by rebuilding the migration inside the entry script.
Phase 2 turns the wrong experiments of the stronger agents mostly into
passes, and those of the weaker agents into judge rejections, fidelity gaps
and freeze violations. GLM-5.3-Flash is the clearest case: its eleven phase-1
crashes become two, because in phase 2 it rewrites the migration inside the
entry script, which the freeze forbids, and six of its cells fail on that
rule instead.

\section{Per-Cell Results}
\label{app:cells}

\begin{figure}[h]
\begin{center}
\includegraphics[width=\linewidth]{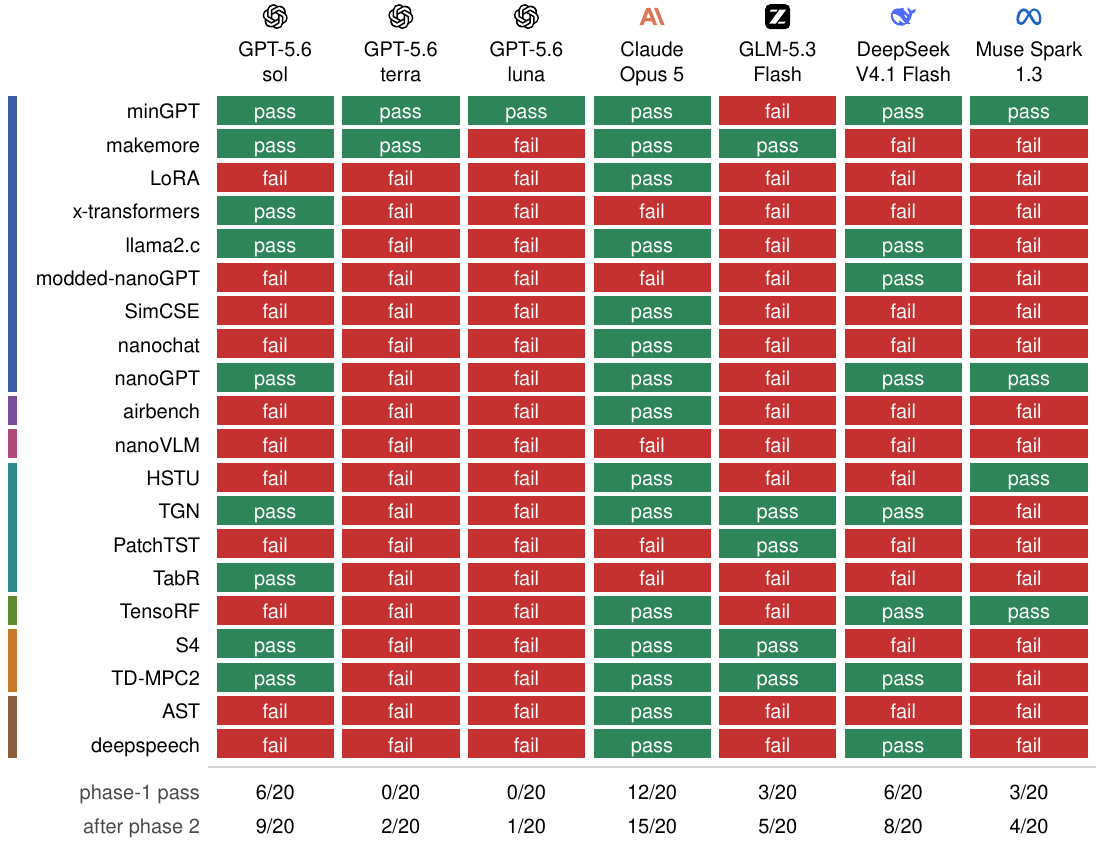}
\end{center}

\caption{Every cell, repositories grouped by
family: the final result, pass or fail.}
\label{fig:matrix}
\end{figure}

Figure~\ref{fig:matrix} gives the result of every cell.
Rows are the \nSlate{} slate repositories and columns the \nAgents{} agents; the
last two rows repeat the per-agent rates of Figure~\ref{fig:accuracy}.
Table~\ref{tab:gates} gives the pass rate per agent after each check.
Claude Opus 5 on PatchTST and GPT-5.6-terra on deepspeech reach the target
and pass the judges, but their exported weights do not load under the
source's parameter names, so the figure draws them as fails.
Two repositories bracket the matrix. minGPT, the smallest source, passes for
every agent but GLM-5.3-Flash; nanoVLM, the longest pipeline, passes for
none. Five repositories (LoRA, SimCSE, nanochat, airbench, AST) pass for
Claude Opus 5 alone, and that is where its lead over the field comes from.
The mid-field agents pass the small language-model repositories (minGPT,
makemore, nanoGPT, llama2.c) and a few others each; GPT-5.6-terra and GPT-5.6-luna
pass little beyond minGPT and makemore. Phase 2 adds between one and three
passes per agent and breaks the three phase-1 ties: GPT-5.6-sol above DeepSeek,
GLM-5.3-Flash above Muse Spark, and GPT-5.6-terra above GPT-5.6-luna.

\section{Cost}
\label{app:cost}

\textbf{Cost and time.} At list prices as of \svPriceDate{} (Table~\ref{tab:agents}), the \nSlate{} cells cost
\svTotalCostLuna{} for luna, \svTotalCostTerra{} for terra,
\svTotalCostGlm{} for GLM-5.3-Flash, \svTotalCostDeepseek{} for DeepSeek,
\svTotalCostMuse{} for Muse Spark, \svTotalCostSol{} for sol and
\svTotalCostClaude{} for Claude Opus 5 (\svCellCostClaude{} per cell, of
which phase 1 is \svCostClaude); \svCacheHitPct{} of input tokens
are cache hits. The list prices of Table~\ref{tab:agents} are all undiscounted, and we used no promotional price when computing costs.

\section{GPU-Only Experiment}
\label{app:earlier}

Before the TPU benchmark we ran a GPU-only generation of the study: twelve
small PyTorch repositories, each migrated whole to JAX by GPT-5.5 at high
reasoning effort and graded on the same GPU by ten checks
(Table~\ref{tab:gpu-pilot}; Section~\ref{sec:taxonomy}). The twelve are
DDPM~\citep{wang2020ddpm}, makemore~\citep{karpathy2022makemore},
minGPT~\citep{karpathy2020mingpt}, nanochat~\citep{karpathy2025nanochat},
nanoGPT~\citep{karpathy2022nanogpt}, nano-vLLM~\citep{yu2025nanovllm},
pyGCN~\citep{kipf2017pygcn}, x-transformers~\citep{wang2020xtransformers},
and the MNIST, REINFORCE, VAE and word language model examples of
PyTorch~\citep{pytorch2016examples}. The checks are
what a careful reviewer of a translation would write: the migration has
its own entry points and no PyTorch, the same weights give the same
logits, loss and gradients, optimizer and data pipeline match,
checkpoints round-trip, greedy decoding matches token for token, and a
restarted run resumes. The agent met 116 of 117, and the single miss
is a convergence band (x-transformers, A6).

\input{figures/table_gpu_pilot.tex}

\section{Code Size and Compute}
\label{app:candidates}

Every cell had enough compute: Figure~\ref{fig:cand-code-and-compute} gives the
TPU chip-hours of all \svCells{} cells, split into the agents' turns and our grading.
Grading, not generation, is where the compute goes: our full-scale grades
took more than twice the chip-hours of the agents' own turns, and the
v6e machines carried nearly three quarters of the total. The table converts
chip-hours to H100-equivalents at each chip's peak bf16 throughput relative
to an H100, so the campaign is comparable with GPU budgets.

\input{figures/fig25_code_and_compute.tex}

The left panel of Figure~\ref{fig:cand-code-and-compute} sets each
passing migration's size against its source: \svPassPortsLarger{} of the
\svPassPortsCounted{} passes are larger than the repository they came from.
Most passing migrations sit within a factor of two of their source's size,
and the agents differ systematically: Claude Opus 5, DeepSeek and GLM write
migrations about as long as the source; GPT-5.6-sol and GPT-5.6-terra write compact
ones, under a thousand lines even for sources ten times that size. The two
largest sources that were migrated successfully, x-transformers and HSTU,
came out at a fraction of their size: both are libraries, and their passing
migrations preserve the model, training and evaluation paths the headline
experiment runs rather than every option the library offers.

\end{document}

%% file: figures/numbers.tex
\newcommand{\svCells}{140}

\newcommand{\svPhaseOneAll}{30/140}
\newcommand{\svPhaseOneAllPct}{21\%}

\newcommand{\svFullAll}{44/140}
\newcommand{\svFullAllPct}{31\%}
\newcommand{\svPhaseOneSol}{6/20}

\newcommand{\svTotalCostSol}{\$323}

\newcommand{\svTotalCostTerra}{\$40}

\newcommand{\svTotalCostLuna}{\$10.0}

\newcommand{\svPhaseOneClaude}{12/20}

\newcommand{\svCostClaude}{\$49.42}
\newcommand{\svCellCostClaude}{\$92.18}
\newcommand{\svTotalCostClaude}{\$1,844}

\newcommand{\svTotalCostGlm}{\$32}

\newcommand{\svPhaseOneDeepseek}{6/20}

\newcommand{\svCrashDeepseek}{1}

\newcommand{\svTotalCostDeepseek}{\$37}

\newcommand{\svTotalCostMuse}{\$283}

\newcommand{\svTaxFailsOne}{110}

\newcommand{\svTaxWrongPctOne}{27\%}
\newcommand{\svTaxCodePctOne}{47\%}

\newcommand{\svTaxFreezeTwo}{16}

\newcommand{\svCostFitSlope}{25}
\newcommand{\svCostFitR}{0.81}
\newcommand{\svCostFitRTwo}{0.66}
\newcommand{\svCostFitP}{0.027}
\newcommand{\svCostFitN}{7}

\newcommand{\svPhaseTwoRan}{108}
\newcommand{\svPhaseTwoRescued}{16}

\newcommand{\svHeadlinePasses}{46}

\newcommand{\svAuditFlagged}{2}

\newcommand{\svGraderPasses}{102}
\newcommand{\svGraderBlind}{58}
\newcommand{\svJudgeFailedBlind}{28}
\newcommand{\svJudgeFailedRescue}{28}

\newcommand{\svJudgeFailed}{56}

\newcommand{\svJudgeBroken}{21}

\newcommand{\svGraderRescue}{44}

\newcommand{\svGateBarDeepseek}{15/20}

\newcommand{\nAgents}{7}

\newcommand{\svTotalTokensB}{12.5}
\newcommand{\svCacheHitPct}{97\%}
\newcommand{\nLabs}{5}

\newcommand{\nSlate}{20}

\newcommand{\svTpuChipHours}{22,802}

\newcommand{\svPriceDate}{September 14, 2026}

\newcommand{\nBarPublished}{15}

\newcommand{\nBarAnchor}{3}

%% file: figures/fig2_fleet_links.tex
\begingroup%
\setlength{\unitlength}{\linewidth}%
\begin{picture}(1,0.33725106)%
\put(0,0){\includegraphics[width=\linewidth]{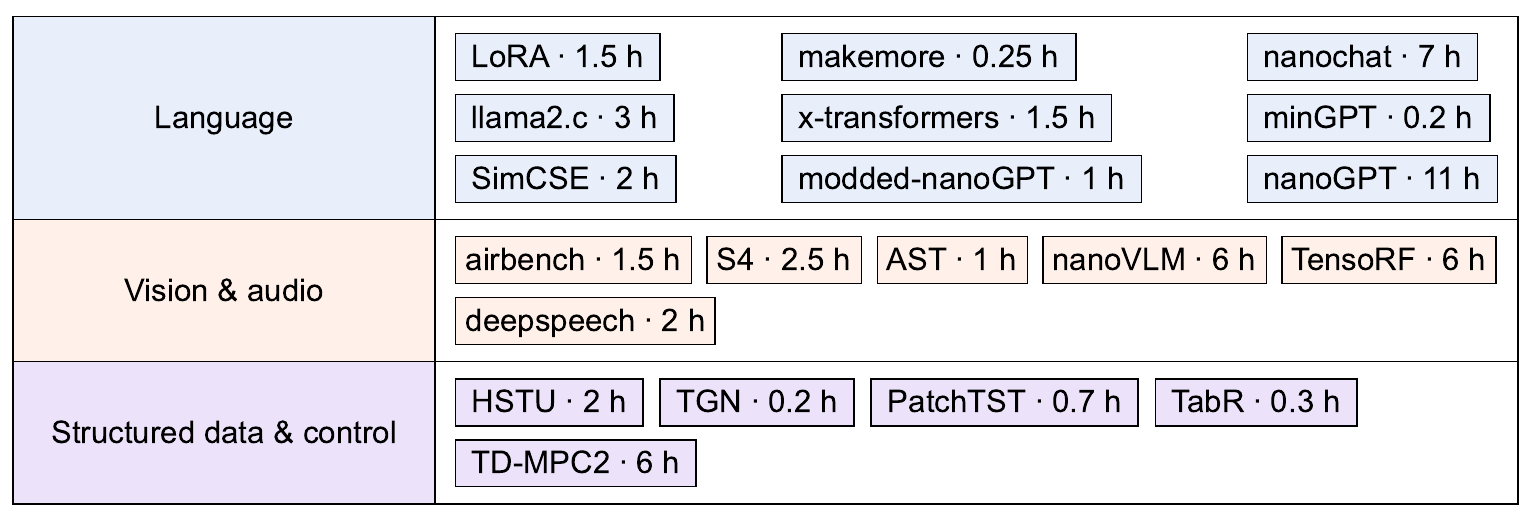}}%
\put(0.30811296,0.28929971){\href{https://github.com/microsoft/LoRA/tree/c4593f060e6a368d7bb5af5273b8e42810cdef90}{\phantom{\rule{0.04999184\unitlength}{0.02244532\unitlength}}}}%
\put(0.52134346,0.28929971){\href{https://github.com/karpathy/makemore/tree/988aa59e4d8fefa526d06f3b453ad116258398d4}{\phantom{\rule{0.09590271\unitlength}{0.02244532\unitlength}}}}%
\put(0.82537545,0.28929971){\href{https://github.com/karpathy/nanochat/tree/dc54a1a3077cab11d68fac4c5d1cd5c51f5d8c7a}{\phantom{\rule{0.08365981\unitlength}{0.02244532\unitlength}}}}%
\put(0.30811296,0.24951028){\href{https://github.com/karpathy/llama2.c/tree/350e04fe35433e6d2941dce5a1f53308f87058eb}{\phantom{\rule{0.07549788\unitlength}{0.02244532\unitlength}}}}%
\put(0.52134346,0.24951028){\href{https://github.com/lucidrains/x-transformers/tree/dfd829b8c1e291ad03c95f509fc874d55f3da40d}{\phantom{\rule{0.13161117\unitlength}{0.02244532\unitlength}}}}%
\put(0.82537545,0.24951028){\href{https://github.com/karpathy/minGPT/tree/37baab71b9abea1b76ab957409a1cc2fbfba8a26}{\phantom{\rule{0.07447764\unitlength}{0.02244532\unitlength}}}}%
\put(0.30811296,0.20972086){\href{https://github.com/princeton-nlp/SimCSE/tree/49fa580a853752ede55b8c76d9debf748e214d3f}{\phantom{\rule{0.07753836\unitlength}{0.02244532\unitlength}}}}%
\put(0.52134346,0.20972086){\href{https://github.com/KellerJordan/modded-nanogpt/tree/1b51e26d304f647c7c12201b3f1513ee5a429ec4}{\phantom{\rule{0.16731962\unitlength}{0.02244532\unitlength}}}}%
\put(0.82537545,0.20972086){\href{https://github.com/karpathy/nanoGPT/tree/3adf61e154c3fe3fca428ad6bc3818b27a3b8291}{\phantom{\rule{0.08672054\unitlength}{0.02244532\unitlength}}}}%
\put(0.30403200,0.15666830){\href{https://github.com/KellerJordan/cifar10-airbench/tree/4c1b6d1e3889b037efadcfd5c0ea65b246592362}{\phantom{\rule{0.07855860\unitlength}{0.02244532\unitlength}}}}%
\put(0.46829089,0.15666830){\href{https://github.com/state-spaces/s4/tree/e757cef57d89e448c413de7325ed5601aceaac13}{\phantom{\rule{0.02448580\unitlength}{0.02244532\unitlength}}}}%
\put(0.57949723,0.15666830){\href{https://github.com/YuanGongND/ast/tree/31088be8a3f6ef96416145c4b8d43c81f99eba7a}{\phantom{\rule{0.03876918\unitlength}{0.02244532\unitlength}}}}%
\put(0.68764284,0.15666830){\href{https://github.com/huggingface/nanoVLM/tree/6ba9082e16f1fc8c21a1f8d0c54b26c9233c8771}{\phantom{\rule{0.08672054\unitlength}{0.02244532\unitlength}}}}%
\put(0.84373980,0.15666830){\href{https://github.com/apchenstu/TensoRF/tree/6a1e23b10b9521221049aec843ab370a747a7982}{\phantom{\rule{0.08161933\unitlength}{0.02244532\unitlength}}}}%
\put(0.30403200,0.11687888){\href{https://github.com/SeanNaren/deepspeech.pytorch/tree/b88a631ef6b96553e16bbaa5bd6a621435bed278}{\phantom{\rule{0.11120633\unitlength}{0.02244532\unitlength}}}}%
\put(0.30811296,0.06382631){\href{https://github.com/meta-recsys/generative-recommenders/tree/fb0ba644be8dd1c7b072199c9c8bc6ce3adaf3d8}{\phantom{\rule{0.05509305\unitlength}{0.02244532\unitlength}}}}%
\put(0.44176461,0.06382631){\href{https://github.com/twitter-research/tgn/tree/d55bbe678acabb9fc3879c408fd1f2e15919667c}{\phantom{\rule{0.04285015\unitlength}{0.02244532\unitlength}}}}%
\put(0.57949723,0.06382631){\href{https://github.com/yuqinie98/PatchTST/tree/204c21efe0b39603ad6e2ca640ef5896646ab1a9}{\phantom{\rule{0.09080150\unitlength}{0.02244532\unitlength}}}}%
\put(0.76518120,0.06382631){\href{https://github.com/yandex-research/tabular-dl-tabr/tree/17baa9082506f8e7a0f8d11bb1e08212926a1507}{\phantom{\rule{0.04795136\unitlength}{0.02244532\unitlength}}}}%
\put(0.30811296,0.02403689){\href{https://github.com/nicklashansen/tdmpc2/tree/e9f59321933cbc8e11a002b842adc7d4ffae8ff1}{\phantom{\rule{0.09080150\unitlength}{0.02244532\unitlength}}}}%
\end{picture}%
\endgroup%

%% file: figures/table_agents.tex
GPT-5.6-sol & Codex CLI & high & 5 / 0.5 / 30 \\
GPT-5.6-terra & Codex CLI & high & 2 / 0.2 / 12 \\
GPT-5.6-luna & Codex CLI & high & 0.2 / 0.02 / 1.2 \\
Claude Opus 5 & Claude Code CLI (Anthropic) & high & 5 / 0.5 / 25 \\
GLM-5.3-Flash & Claude Code CLI (Z.ai) & high & 0.15 / 0.03 / 0.5 \\
DeepSeek V4.1 Flash & Claude Code CLI (DeepSeek) & high & 0.3 / 0.006 / 1.2 \\
Muse Spark 1.3 & Claude Code CLI (Meta) & high & 1.25 / 0.15 / 4.25 \\
\bottomrule

%% file: figures/table_checks_cost.tex
\begin{tabular}{l cccc ccccc}
\toprule
 & \multicolumn{4}{c}{pass rate} & \multicolumn{5}{c}{cost and time} \\
\cmidrule(lr){2-5} \cmidrule(l){6-10}
agent & target & +correctness & +cheat & +weights & pass & tokens & \$/cell & \$/pass & hours \\
\midrule
Claude-Opus-5 & \textbf{95\%} & \textbf{80\%} & \textbf{80\%} & \textbf{75\%} & \textbf{15/20} & \textbf{156M} & \textbf{92} & \textbf{123} & 8.7 \\
GPT-5.6-sol & 75\% & 70\% & 45\% & 45\% & 9/20 & 23M & 16 & 36 & 1.0 \\
DS-4.1-Flash & 75\% & 60\% & 40\% & 40\% & 8/20 & 106M & 1.9 & 4.7 & 11.1 \\
GLM-5.3-Flash & 60\% & 50\% & 25\% & 25\% & 5/20 & 43M & 1.6 & 6.3 & \textbf{12.6} \\
Muse-Spark-1.3 & 50\% & 35\% & 20\% & 20\% & 4/20 & 32M & 14 & 71 & 4.1 \\
GPT-5.6-terra & 35\% & 25\% & 15\% & 10\% & 2/20 & 5.3M & 2.0 & 20 & 0.4 \\
GPT-5.6-luna & 20\% & 15\% & 5\% & 5\% & 1/20 & 16M & 0.5 & 10.0 & 0.6 \\
\midrule
all & 59\% & 48\% & 33\% & 31\% & 44/140 & 54M & 18 & 58 & 5.5 \\
\bottomrule
\end{tabular}

%% file: figures/table_fleet.tex
\makebox[0pt][l]{\textbf{Language}} & & & & \\*
llama2.c & language modeling & a best validation loss of at most 1.130 at the source's own 100,000-iteration default horizon & $\le$ 1.13 & anchor \\
LoRA & fine-tuning & a corpus BLEU of at least 0.680 on the E2E test set & $[0.68, 1.0]$ & readme \\
makemore & language modeling & a best test NLL of at most 2.023 & $\le$ 2.023 & anchor \\
minGPT & language modeling & at least 495 of 500 correct adder test answers & $\ge$ 495 & rule \\
modded-nanoGPT & language modeling & a FineWeb validation cross-entropy loss of at most 3.285 & $\le$ 3.285 & readme \\
nanochat & language modeling & a base-evaluation-suite score of at least 0.256525 & $[0.2565, 1.0]$ & readme \\
nanoGPT & language modeling & a validation loss at the graded iteration budget & $\le$ 3.12 & readme \\
SimCSE & sentence embeddings & an average Spearman correlation of at least 73.5 across the seven STS test sets & $[73.5, 100.0]$ & readme \\
x-transformers & language modeling & a validation loss between 0.8788 and 0.9788 & $[0.8788, 0.9788]$ & anchor \\
\addlinespace[3pt]
\makebox[0pt][l]{\textbf{Vision}} & & & & \\*
airbench & image classification & a test-time-augmented CIFAR-10 test accuracy of at least 0.9393, taken as the mean over the source's own 25-trial loop & $[0.9393, 1.0]$ & paper \\
\addlinespace[3pt]
\makebox[0pt][l]{\textbf{Generative \& multimodal}} & & & & \\*
nanoVLM & vision-language & a best MMStar accuracy of at least 0.353 & $[0.353, 1.0]$ & readme \\
\addlinespace[3pt]
\makebox[0pt][l]{\textbf{Structured data}} & & & & \\*
HSTU & recommendation & a MovieLens-20M HR@10 of at least 0.3450 & $[0.345, 1.0]$ & readme \\
PatchTST & time-series forecasting & a test mean squared error of at most 0.380 in standardized units & $\le$ 0.38 & paper \\
TabR & tabular learning & a test accuracy of at least 0.9745 on the Covertype test split of 116,203 rows & $[0.9745, 1.0]$ & paper \\
TGN & temporal graph learning & a transductive test average precision of at least 0.980 for future edge prediction on JODIE Wikipedia & $[0.98, 1.0]$ & paper \\
\addlinespace[3pt]
\makebox[0pt][l]{\textbf{Science \& 3D}} & & & & \\*
TensoRF & neural fields & a mean test-view PSNR of at least 36.10 dB on Synthetic-NeRF lego & $[36.1, 45.0]$ & paper \\
\addlinespace[3pt]
\makebox[0pt][l]{\textbf{Control \& sequence}} & & & & \\*
S4 & long-sequence modeling & a top-1 accuracy of at least 87.0 percent on a held-out CIFAR-10 split & $[87.0, 100.0]$ & readme \\
TD-MPC2 & reinforcement learning & a mean evaluation return of at least 750.0 over the source's 10-episode evaluation on DMControl walker-run & $[750.0, 1000.0]$ & readme \\
\addlinespace[3pt]
\makebox[0pt][l]{\textbf{Audio \& speech}} & & & & \\*
AST & audio classification & a test-split top-1 accuracy of at least 0.975 & $[0.975, 1.0]$ & readme \\
deepspeech & speech recognition & a word error rate of at most 20.0 percent on the AN4 test set & $[1.5, 20.0]$ & released-checkpoint \\

%% file: figures/table_gpu_pilot.tex
\begin{table}[!ht]
\centering
\begin{minipage}[t]{0.35\linewidth}\vspace{0pt}\centering
\begin{tabular}{@{}lc@{}}
\toprule
Repository & Checks \\
\midrule
DDPM & 10/10 \\
makemore & 10/10 \\
minGPT & 10/10 \\
MNIST & 9/9 \\
nano-vLLM & 9/9 \\
nanochat & 10/10 \\
nanoGPT & 10/10 \\
pyGCN & 9/9 \\
REINFORCE & 10/10 \\
VAE & 10/10 \\
word language model & 10/10 \\
x-transformers & 9/10$^{\,\text{A6}}$ \\
\midrule
All & 116/117 \\
\bottomrule
\end{tabular}
\end{minipage}\hfill
\begin{minipage}[t]{0.61\linewidth}\vspace{0pt}\centering
\begin{tabular}{@{}l@{\hspace{5pt}}l@{\hspace{7pt}}p{0.60\linewidth}@{}}
\toprule
 & Check & Passes if the migration has \ldots \\
\midrule
A0 & Completeness & a JAX code base with its own entry points and no PyTorch inside \\
A1 & Forward parity & same weights give the same logits and loss as PyTorch \\
A2 & Gradient parity & same gradients as PyTorch autograd \\
A3 & Optimizer & the learning-rate schedule and optimizer step match the source \\
A4 & Data pipeline & batches are the source's token windows \\
A5 & Checkpoints & weights round-trip between the two frameworks \\
A6 & Convergence & its own trainer reaches the source's validation-loss band \\
A7 & Resume & a restarted training run continues from its checkpoint \\
A8 & Generation & greedy decoding matches the source token for token \\
A9 & Runnability & its own train and sample scripts run stand-alone \\
\bottomrule
\end{tabular}
\end{minipage}
\caption{\textbf{The GPU-only experiment.} Left: checks passed per repository by GPT-5.5 at high reasoning effort, whole-repository PyTorch to JAX migrations graded on an A100 against the running source (the GPU generation's last run); 11 of 12 repositories passed every check (x-transformers missed A6). Right: the ten checks, each a test against the running PyTorch source. MNIST and pyGCN have no generation and nano-vLLM no training resume, so they have nine. Five of the twelve repositories are in \bench.}
\label{tab:gpu-pilot}
\end{table}

%% file: figures/fig25_code_and_compute.tex
\providecommand{\svPassPortsCounted}{44}
\providecommand{\svPassPortsLarger}{16}
\begin{figure}[!ht]
\centering
\makebox[\linewidth][s]{%
\raisebox{-26.02pt}{\includegraphics[width=295.04pt]{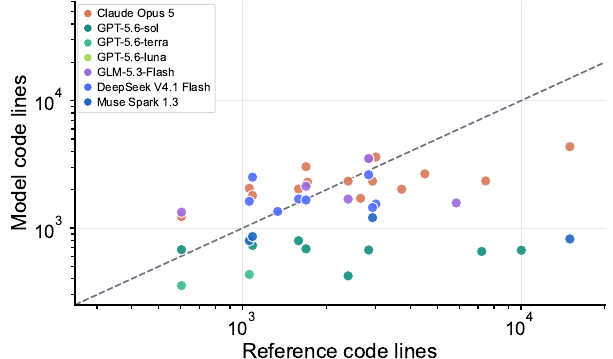}}\hfill
\raisebox{\depth}{\setlength{\tabcolsep}{4pt}\input{figures/table_compute.tex}}%
}
\caption{\textbf{Code size and compute.} Left: Python lines of each passing migration against
the source it came from, one mark per cell (\svPassPortsCounted{} cells; the dashed line is equal size). Right: TPU chip-hours of all cells, split
into the agents' turns (generation) and our grading of their code, and the same in H100-equivalent hours at peak
bf16 throughput.}
\label{fig:cand-code-and-compute}
\end{figure}

%% file: figures/table_compute.tex
\begin{tabular}{@{}lrrr@{}}
\toprule
 & Generation & Grading & Total \\
\midrule
\multicolumn{4}{@{}l}{TPU chip-hours} \\
\quad v4 & 71 & 455 & 526 \\
\quad v5p & 633 & 5,163 & 5,796 \\
\quad v6e & 5,940 & 10,540 & 16,480 \\
\quad all & 6,644 & 16,158 & 22,802 \\
\addlinespace[3pt]
\multicolumn{4}{@{}l}{H100-equivalent hours} \\
\quad v4 & 20 & 126 & 146 \\
\quad v5p & 294 & 2,396 & 2,690 \\
\quad v6e & 5,514 & 9,782 & 15,296 \\
\quad all & 5,828 & 12,304 & 18,132 \\
\bottomrule
\end{tabular}